\documentclass{article} 
\usepackage{iclr2024_conference,times}
\usepackage[pdftex]{graphicx}
\usepackage{graphbox}
\usepackage{wrapfig}
\usepackage{listings}
\usepackage{subcaption}
\usepackage{bigstrut}
\usepackage{algorithm}
\usepackage{algpseudocode}
\usepackage{pgfplots}
\usepackage{tikz}
\usepackage{tikz-3dplot}
\usetikzlibrary{math}
\usetikzlibrary{decorations.text}
\usepackage{amssymb}
\usepackage{pifont}
\newcommand{\cmark}{\color{green}{\ding{51}}}%
\newcommand{\xmark}{\color{red}\ding{55}}%
\usepackage[export]{adjustbox}
\usepackage{enumitem}
\usepackage{float}

\usepackage{amsmath,amsfonts,bm}

\def\eqref#1{equation~\ref{#1}}

\def\1{\bm{1}}

\DeclareMathAlphabet{\mathsfit}{\encodingdefault}{\sfdefault}{m}{sl}
\SetMathAlphabet{\mathsfit}{bold}{\encodingdefault}{\sfdefault}{bx}{n}

\newcommand{\R}{\mathbb{R}}

\usepackage{hyperref}
\usepackage{url}

\title{On the Surprising Efficacy of Distillation as an Alternative to Pre-Training Small Models}

\author{Sean Farhat \& Deming Chen \\
Department of Computer Science\\
University of Illinois, Urbana-Champaign \\
\texttt{\{seanf2, dchen\}@illinois.edu}
}

\iclrfinalcopy 
\begin{document}

\maketitle

\begin{abstract}
   In this paper, we propose that small models may not need to absorb the cost of pre-training to reap its benefits. 
   Instead, they can capitalize on the astonishing results achieved by modern, enormous models to a surprising degree. 
   We observe that, when distilled on a task from a pre-trained teacher model, a small model can achieve or surpass the performance it would achieve if it was pre-trained then finetuned on that task.  
   To allow this phenomenon to be easily leveraged, we establish a connection reducing knowledge distillation to modern contrastive learning, opening two doors: (1) vastly different model architecture pairings can work for the distillation, and (2) most contrastive learning algorithms rooted in the theory of Noise Contrastive Estimation can be easily applied and used.
   To illustrate these points, we demonstrate this paradigm using pre-trained teacher models from open-source model hubs, Transformer and convolution based model combinations, and a novel distillation algorithm that massages the Alignment/Uniformity perspective of contrastive learning by \citet{wang2020align_uniform} into a distillation objective. We choose this flavor of contrastive learning due to its low computational cost, an overarching theme of this work.
   We also observe that this phenomenon tends not to occur if the task is data-limited. However, this can be alleviated by leveraging yet another scale-inspired development: large, pre-trained generative models for dataset augmentation. Again, we use an open-source model, and our rudimentary prompts are sufficient to boost the small model's performance.
   Thus, we highlight a training method for small models that is up to 94\% faster than the standard pre-training paradigm without sacrificing performance.
   For practitioners discouraged from fully utilizing modern foundation datasets for their small models due to the prohibitive scale, we believe our work keeps that door open.
\end{abstract}

\section{Introduction}
\begin{wrapfigure}[20]{R}{.5\textwidth}
\vspace{-14.5pt}
\tdplotsetmaincoords{70}{120}
\begin{tikzpicture}[tdplot_main_coords,scale=1.5]
	\tikzmath{function funcion(\x,\y) {return 1.75+0.25*sin((0.5*\x + \y) r);};}
	\pgfmathsetmacro{\step}{pi/50.0}
	\pgfmathsetmacro{\xf}{1.0*pi}
	\pgfmathsetmacro{\yf}{1.0*pi}
	\pgfmathsetmacro{\h}{1.5}
	\pgfmathsetmacro{\a}{0.5}
	\pgfmathsetmacro{\b}{\a+1.5}
	\pgfmathsetmacro{\c}{1.0}
	\pgfmathsetmacro{\d}{\c+1.5}
	\pgfmathsetmacro{\e}{\b+.5}
    \pgfmathsetmacro{\zA}{funcion(\a,\c)}
	\pgfmathsetmacro{\zB}{funcion(\b,\c)}
	\pgfmathsetmacro{\zC}{funcion(\b,\d)}
	\pgfmathsetmacro{\zD}{funcion(\a,\d)}
	\pgfmathsetmacro{\zE}{funcion(\e,\d)}
	\pgfmathsetmacro{\zF}{funcion(\e,\c)}
	\draw[thick,->] (0,0,0) -- (\xf,0,0) node [below] {$x$}; 
	\draw[thick,->] (0,0,0) -- (0,\yf,0) node [right] {$y$}; 
	\draw[thick,->] (0,0,0) -- (0,0,\h+0.5,0) node [above] {$z$}; 
	\draw[white] (\a,\d,0) -- (\b,\d,0) node [black,below,sloped,midway] {Task};
	\fill[gray!25] (\a,\c,0) -- (\b,\c,0) -- (\b,\d,0) -- (\a,\d,0) -- (\a,\c,0);
    \draw[densely dashed] (\b,\c,0) -- (\b,\d,0);
	\draw[white] (\e,\c,0) -- (\e,\d,0) node [black,below,sloped,midway] {Synthetic};
	\fill[red!25] (\b,\c,0) -- (\e,\c,0) -- (\e,\d,0) -- (\b,\d,0) -- (\b,\c,0);
	\draw[dash dot dot] (\a,\c,0) -- (\a,\c,\zA);
	\draw[dash dot dot] (\b,\c,0) -- (\b,\c,\zB);
	\draw[dash dot dot] (\a,\d,0) -- (\a,\d,\zD);
	\draw[dash dot dot] (\b,\d,0) -- (\b,\d,\zC);
	\draw[dash dot dot] (\e,\d,0) -- (\e,\d,\zE);
	\draw[dash dot dot] (\e,\c,0) -- (\e,\c,\zF);
    \draw[cyan,opacity=0.5,
        samples=15,
        postaction={decorate, 
            decoration={text along path, raise=2pt, text={|\small|Pre-trained manifold},
            text align={align=center}}} 
    ]
    plot[domain=0:\yf,smooth,variable=\t] ({0},{\t},{funcion(0,\t)});
	\foreach \x in {0,\step,...,\xf}{
        \draw[cyan,opacity=0.5]
        plot[domain=0:\yf,smooth,variable=\t] ({\x},{\t},{funcion(\x,\t)});
	}

	\foreach \y in {0,\step,...,\yf}{
		\draw[cyan,opacity=0.5] plot[domain=0:\yf,smooth,variable=\t] ({\t},{\y},{funcion(\t,\y)});
	}
	\foreach \x in {\a,\e}
        \draw[blue,thick,opacity=0.85] plot[domain=\c:\d,smooth,variable=\t] ({\x},{\t},{funcion(\x,\t)});

    \draw[blue,densely dashed,opacity=0.85] plot[domain=\c:\d,smooth,variable=\t] ({\b},{\t},{funcion(\b,\t)});

    \foreach \y in {\c,\d}
        \draw[blue,thick,opacity=0.85] plot[domain=\a:\e,smooth,variable=\t] ({\t},{\y},{funcion(\t,\y)});
\end{tikzpicture}
\caption{By only mimicking the relevant slice of a pre-trained manifold, 
a small model can achieve the \emph{same or better} performance
than if it had been fully pre-trained and finetuned.
Adding synthetic samples leads to better generalization.}
\label{fig:slice}
\end{wrapfigure}
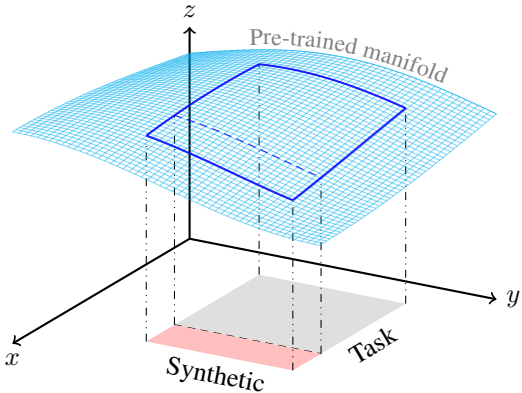

Small machine learning models are incredibly valuable: cost savings on 
memory and compute, reductions on energy footprints, and deployment on resource-constrained edge devices.
However, recent attention has focused on training immense models on immense datasets, dubbed ``foundation models''
\citep{radford2021clip, radford2019gpt2, brown2020gpt3, chen2020simclr, bommasani2021foundations, alayrac2022flamingo, yuan2021florence}.
Their surprisingly high performance and generalization capabilities has led to the current motivating sentiment of unimpeded scaling. How can small models keep up?

Various lines of research tackle how to squeeze out maximal performance from small models, all 
targeting different points in the training process.
Before training, neural architecture search (NAS) \citep{zoph2017nas,liu2019darts} 
algorithmically constructs small yet high-performing model designs.
After training, one can prune or quantize \citep{jacob2017quantization,han2016pruning} a model to
reduce its size with a minimal performance drop.
We focus on what happens in the middle: the training process.
During training, knowledge distillation (KD) \citet{caruana2006compression,hinton2015distilling} can 
assist the model by having it also learn from a larger teacher model that is an expert at the task.

We consider the situation of taking the main strategy that defines foundation models' success, pre-training, and applying it to small models. This has been a classic strategy for boosting the performance of models in general, however the modern scale of foundation datasets growing to billions of samples \citet{schuhmann2022laion} puts pre-training at odds with the main appeal of small models: low cost at both train and inference time. 
One can hope that a pre-trained small model exists in a public model hub, however what if a custom model architecture is desired? Then, a choice must be made:
either (a) absorb the cost of pre-training and reap the performance benefits, or (b) train from scratch on just the desired task, absorbing the probable reduction in performance that comes with decreasing model size.
Addressing this dilemma, we re-evaluate if pre-training is necessary for small models. Often, the demand for small models stems from deploying them for one or a handful of tasks. Therefore, does a small model need a \emph{comprehensive} feature backbone?
Alternately, what if we teach it to behave like it was pre-trained and finetuned, but only on the relevant slice of knowledge (see Figure
\ref{fig:slice})?

In this paper, we show that, by leveraging the progress in foundation models, contrastive learning,
and pre-trained generative models, small models can achieve
and \emph{surpass} pretrain-then-finetune performance without ever needing to touch a pre-training dataset.
Our approach is simple yet effective: 
(1) using \emph{any} publicly available teacher pre-trained on an appropriate foundation dataset, optionally finetune it for the desired task, then
(2) transfer its knowledge of the task via our proposed distillation loss on 
(3) the task dataset augmented with synthetic samples generated from publicly available pre-trained generative models.
Our method can easily be applied to improve performance for any small model on any task, can be used in continual learning
regimes \citep{li2017lwf} if multiple tasks are desired, and fits in nicely between methods like NAS and pruning
to fully utilize all the tools to maximize small model accuracy.

We test our approach on 6 visual recognition tasks, spanning both the data-abundant and data-limited regimes, the latter of which benefits the most from pre-training.
The formulation of our KD loss as a contrastive objective allows our setup to be agnostic to the underlying teacher-student architectures, a luxury rarely granted from prior
work on distillation algorithms.
We demonstrate this by utilizing teachers representative of the main vision architectures, a Vision Transformer \citep{dosovitskiy2020vit} and 50-layer ResNet \citep{he2016resnet}, 
to assist 2 small students, a MobileNetV2 \citep{sandler2018mobilenetv2} and 18-layer ResNet.
We also provide a cost analysis of our method: skipping pre-training leads to large resource benefits by reducing overall training time. However, a nontrivial image generation cost is added given 
the state of the diffusion model we employ.
Our code can be found at \url{https://github.com/sfarhat/dapt}.

\section{Related Work}
\paragraph{Knowledge Distillation}
\label{sec:background_kd}
Knowledge Distillation (KD) first appeared in the realm of ensembling methods \citep{Dietterich2000EnsembleMI}, 
though it was \citet{caruana2006compression} who first used $n \geq 1$ expert models to teach a smaller model, thus considering their technique a form of model compression. 
\citet{hinton2015distilling} popularized the method by using a temperature-based softmax. 
Since then, KD has blossomed into an active area of research.
Logit-Based KD algorithms \citep{hinton2015distilling, zhao2022dkd, yang2021srrl, chen2022simkd} look at the logits, the input to the classification softmax. 
Feature-based algorithms \citep{romero2014fitnets,zagoruyko2016at,ahn2019vid, passalis2018pkt, heo2019ab, kim2018ft, huang2017nst, heo2019ofd,
chen2021semckd, chen2021reviewkd, miles2021itrd, srinivas2018jacobian, tian2019crd}
probe the middle layers of both networks and take different perspectives on how intermediate knowledge should be quantified and transferred.
Relation-based methods 
\citep{park2019rkd, liu2021sskd, fsp, tung2019sp, peng2019cc} look at the inter-batch relationships between outputs
of the teacher and student networks.
\citet{tian2019crd} were the first to apply the idea of contrastive learning to KD,
via a formulation inspired by \citet{wu2018unsupervised}, though their work pre-dated most modern advancements in contrastive learning.
The target performance we aim to beat, a pre-trained student, has rarely been addressed. Most prior work looked at distilling
all of ImageNet \citep{imagenet,russakovsky2015ilsvrc} and finetuning the student after to test task performance.

\paragraph{Contrastive Learning}
\label{sec:background_contrastive}
Contrastive Learning comes from the idea of Noise Contrastive Estimation (NCE) \citep{pmlr-v9-gutmann10_nce}. Broadly speaking,
the main idea is to pull ``positive pairs'' together, and push ``negative pairs'' far apart.
Early iterations of this idea 
\citep{contrastive_loss, schroff2015triplet, NIPS2016_6b180037_npairs, pmlr-v2-salakhutdinov07_softnn,frosst2019_softnn2,
oord2018representation_infonce}
dealt with different sources and quantities for positive and negative pairs.
Most modern advancements in contrastive learning come from the realm of self-supervised learning \citep{chen2020simclr, he2020moco, chen2020mocov2, grill2020byol, chen2021simsiam, zbontar2021barlow}.

\paragraph{Generative Models as Data}

Due to the increased realism of synthetic images, recent work has examined if they are good enough to use as a source of data since we can
infinitely sample from generative models. 
Currently, diffusion models
\citep{sohl2015deep, 
ho2020ddpm}, equivalent to score-based methods \citep{JMLR:v6:hyvarinen05-scorematching, denoising_scorematching, song2019ncsn,
song2020score}
offer the best generative guarantees in terms of sampling diversity and likelihood maximization.
The only downside to diffusion models, as of writing, is their slow sampling speed. 
However, improving sampling speeds with minimal effect on sample quality is 
an active area of research \citep{song2023consistency,salimans2022progressive}.
The works of \citet{he2022synthetic,zhou2023training,bansal2023leaving,
sariyildiz2022fake,azizi2023synthetic} used diffusion to augment the training dataset for supervised training with positive results,
but did not consider the KD setting.

\section{Method}
Our main idea is to assist a small model's supervised learning process for task $\mathcal{T}$ by having it 
match the behavior of a large ``teacher'' model $f_T$, which has been pre-trained then finetuned for $\mathcal{T}$.
In doing so, the target small ``student'' model $f_S$ behaves as if it too was pre-trained and finetuned on $\mathcal{T}$, when in reality, it never touches the pre-training dataset.
Our approach builds on the standard KD paradigm by 
(1) deliberately using a pre-trained $f_T$, 
(2) introducing a simple, inexpensive, and architecture-agnostic distillation algorithm deeply linked to contrastive learning, and 
(3) using supplementary synthetic data as needed.

\begin{figure}
    \centering
    \includegraphics[width=.49\textwidth,valign=c]{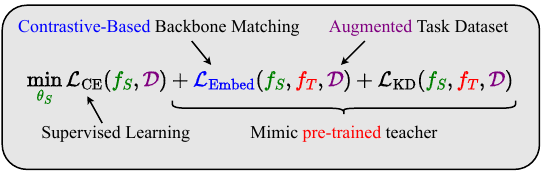}
    \begin{minipage}{0.49\textwidth}
        \centering
        \begin{tabular}[h]{|c|c|c|}
           \hline
           Method & High Accuracy & Low Cost \bigstrut[c] \\
           \hline
           Supervised & \xmark & \cmark \bigstrut[t] \\
           Pre-training & \cmark & \xmark \\
           Ours & \cmark & \cmark \bigstrut[b] \\
           \hline
        \end{tabular}
    \end{minipage}
    \caption{(Left) Our proposed alternative to pre-training and finetuning. (Right) The benefits of our method compared to standard approaches.}
    \label{fig:summary}
\end{figure}

\subsection{Using a Pre-Trained Teacher $f_T$}

In order for $f_S$ to behave like it was pre-trained, it should learn from a teacher $f_T$ that was. 
Model hubs make these teachers easily accessible.
While not necessary, we can additionally finetune $f_T$ on $\tau$ to add an extra learning signal for the student.
To reflect modern 
practices, when we refer to finetuning, we imply the method of \emph{linear probing} (LP) which freezes the backbone weights and 
only updates the task-specific head. 

\subsection{The Contrastive-Based Embedding Loss}

We want to
ensure the student's backbone mimics a pre-trained one on the desired task.
To achieve this and avoid the complexities induced by the implicit biases of model architectures, a common pitfall in prior work,
we look to contrastive learning theory,
which only utilizes the final embeddings $h$.


\paragraph{Contrastive-Based Distillation}

For the sake of simplicity, in this section we will ignore the classification heads and view the student $f_S$ and teacher $f_T$ 
as models that generate high-dimensional representations $h_S \in \R^S, h_T \in \R^T, S \neq T$ after their final backbone layer
(Figure \ref{fig:kd_flow}). 
We believe a robust measure of distance can be effective given its recent success in the self-supervised 
domain: contrastive learning. Following the ideas from their literature, we append ``projection'' modules $g_S: \R^S \to \R^d,
g_T: \R^T \to \R^d$, where $d \ll S, T$. Thus, given a data distribution $x \sim p_{\text{data}}$,
our models induce distributions $p_S(\cdot), p_T(\cdot)$, such that $g_S \circ f_S(x) =: z_S \sim p_S, 
g_T\circ f_T(x) =: z_T \sim p_T$. Note that both $g(\cdot)$'s include a final normalization step that ensures $||z||_2 = 1$.
We also define the distribution of ``positive'' pairs $p_{\text{pos}}(\cdot, \cdot)$, where the marginals should match: 
$\forall z_S, \int p_{\text{pos}}(z_S, z_T)dz_T = p_{S}(z_S)$ and 
$\forall z_T, \int p_{\text{pos}}(z_S, z_T)dz_S = p_{T}(z_T)$.
We approximate sampling from
$p_{\text{pos}}(\cdot, \cdot)$ by passing the same sample $x_i \sim p_{\text{data}}$ through both networks and projectors to 
generate the positive 
pair $(z_{S, i}, z_{T, i})$.
\emph{Any} sample $x_j, j \neq i$ that creates a $z_{\cdot, j}$ induces negative pairs
$(z_{S, i}, z_{T, j}), (z_{S, j}, z_{T, i})$. 

At its core, contrastive learning pulls positive pairs together and pushes negative pairs apart. 
Our proposed formulation that
achieves this, inspired by the InfoNCE loss of \citet{oord2018representation_infonce}, is to minimize the following
with respect to the parameters of the student $f_S$ and both projectors $g_S, g_T$:

\begin{equation}
    \label{eq:kd_contrastive}
    \mathbb{E}_{(z_{S,i}, z_{T,i})\sim p_\text{pos}, \{z_{S,j}\}_{j=1}^M \sim p_{S}, \{z_{T,j}\}_{j=1}^M \sim p_{T}} 
    \left[ - \log \frac{e^{{z_{S, i}}^\top z_{T, i} / \tau}}
            {\sum\limits_{j} e^{{z_{S, i}}^\top z_{T, j} / \tau} + 
            \sum\limits_{j} e^{{z_{S, j}}^\top z_{T, i} / \tau}} \right]
\end{equation}

where $\tau$ is a temperature hyperparameter and $M$ is the number of negative samples chosen beforehand.
This formulation allows us to reduce any contrastive-based objective to a distillation one.
We choose one that is inexpensive and interpretable as an illustration, leaving other contrastive-distillation
collaborations to future work.

\paragraph{Optimizing Alignment and Uniformity}
\citet{wang2020align_uniform} showed 
that as the number of negative samples 
$M \to \infty$, then 2 simpler quantities can be optimized instead: the \emph{alignment} (cosine similarity)
of the positive pairs, and the \emph{uniformity} of all (normalized) samples on the hypershpere in $\R^d$.
With no negative sample bank, momentum model, or large batch size,
this method is the most lightweight contrastive algorithm. Thus, we chose Alignment/Uniformity (\textbf{A/U}) as
the core of our contrastive-distillation algorithm.
We re-express their metrics to reflect our distillation flavor of generating positive pairs:

\begin{align}
    \label{eq:au}
    \mathcal{L}_{\text{align}} := \mathbb{E}_{(z_{S, i}, z_{T, i}) \sim p_{\text{pos}}} [|| z_{S, i} - z_{T, i} ||_2^\alpha] && 
    \mathcal{L}_{\text{uniform}} := \log \mathbb{E}_{(z_{Z, i}, z_{Z, j}) \sim p_{Z}} [e^{-t ||z_{Z, i} - z_{Z, j}||_2^2}] 
\end{align}

where $\alpha > 0, t > 0, Z \in \{S, T\}$.
We can combine these into one term:
$\mathcal{L}_{\text{Embed}} := w_{\text{align}} \cdot \mathcal{L}_{\text{align}} + 
w_{\text{uniform}} \cdot \mathcal{L}_{\text{uniform}}. 
$
and use the default parameters suggested in \citet{wang2020align_uniform}: $w_{\text{align}} = 1, w_{\text{uniform}} = 1, \alpha=2, 
t=2$. Pseudocode can be found in Figure \ref{fig:kd_code}. 

\subsubsection{The Final Training Objective}

If we finetune $f_T$, we can leverage the original method of \citet{hinton2015distilling} and add their effective logit-based loss $\mathcal{L}_{KD}$ (Equation \ref{eq:kd}). Altogether, we can combine the above loss functions with the standard cross-entropy loss $\mathcal{L}_{\text{CE}}$
on the true one-hot labels:
$\mathcal{L} := \lambda_1 \mathcal{L}_{\text{CE}} + \lambda_2 \mathcal{L}_{\text{Embed}} + \lambda_3 \mathcal{L}_{\text{KD}}$.
We used $\lambda_1 = \lambda_2 = \lambda_3 = 1$.
Ablations for the overall A/U design can be found in Appendix \ref{sec:ablations}. 

\subsection{Augmenting the Transfer Dataset}

Lastly, we address the transfer dataset $\mathcal{D}$. Due to the high costs, we avoid touching the pre-training dataset, but \emph{only} using the task dataset is suboptimal. We cannot deny the power of more data, so we propose to apply the known strategy of data augmentation to the distillation setting.
Given the recent successes of diffusion models in generative modelling, we choose to use a pre-trained text-to-image model, 
Stable Diffusion \citep{rombach2022ldm}, as our source of extra samples. 
We leverage the publicly available stable-diffusion-v1-4\footnote{\url{https://huggingface.co/CompVis/stable-diffusion-v1-4}}, which was pre-trained on LAION-2B \citep{schuhmann2022laion}, 
and guide its generation with the appropriate language prompts (Appendix \ref{sec:syn_details}).

\section{Experiments}
\paragraph{Setup}
We choose data-limited datasets since they are most helped by pre-training: 
MIT-67 \citep{quattoni2009mit_indoor}, 
CUB-2011 \citep{Wah2011cub_2011}, DTD \citep{cimpoi14dtd}, 
and Caltech-101 \citep{li2003caltech}. 
In addition, we evaluate on standard vision benchmarks: CIFAR-10 and CIFAR-100 \cite{krizhevsky2009cifar}. 
We test our method with 2 teachers: a ResNet50 \citep{he2016resnet} and a base Vision 
Transformer \citep{dosovitskiy2020vit}, ViT-B-16. For the target small models, we choose a ResNet18 and a 
MobileNetV2 \citep{sandler2018mobilenetv2}.
Details about optimizers, schedulers, and input augmentations can be found in Appendix \ref{sec:experiment_details}.

\paragraph{Results}
We keep our main observations in this section and provide more detailed experiments in Appendices \ref{sec:self-supervised-transfer}, \ref{sec:acc_comparison}, and \ref{sec:additional_pairs}.
The baselines we compare to are when the small model was either (1) pre-trained on ImageNet for 100 epochs and linearly probed for the task or (2) trained on the task from scratch. We denote augmenting the transfer dataset with 
synthetic samples as ``A/U ($n \times$)'', where $n \times$ indicates the number of extra samples in terms of the size of the 
respective train set (see Appendix \ref{sec:dataset_details}). We only use synthetic data in the case where the transfer dataset is so limited that A/U ($0 \times$) lags too far behind the pre-trained goal. 
In addition, we apply standard image augmentations, e.g. manipulating color properties, cropping, and flipping.
The results of the ResNet50-MobileNetV2 pair can be found in 
Figure \ref{fig:costs}. Other pairings can be found in Appendix \ref{sec:additional_pairs}.
The cost of A/U ($n \times$) includes the times to finetune the teacher
and generate the extra images.
As we can see, while maintaining a competitive or superior accuracy to their pre-trained counterpart, our method can cut training time by up to 94\%, though gets slower if more samples need to be generated.
We do not include the time to pre-train the teacher or generative models since those are assumed to be 
obtained already trained off-the-shelf.

\begin{figure}[h]
    \begin{center}
       \includegraphics[width=\textwidth]{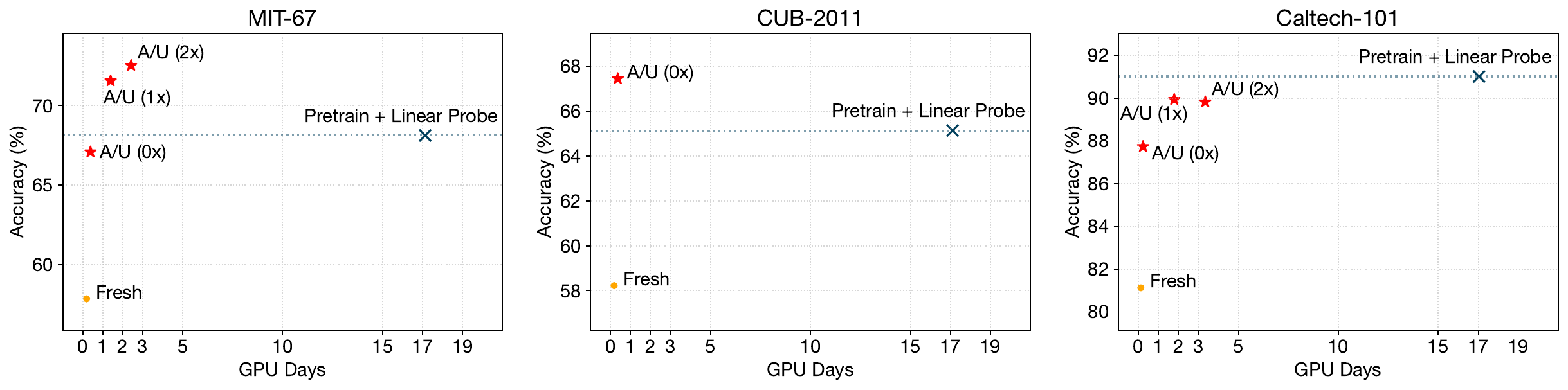} 
       \includegraphics[width=\textwidth]{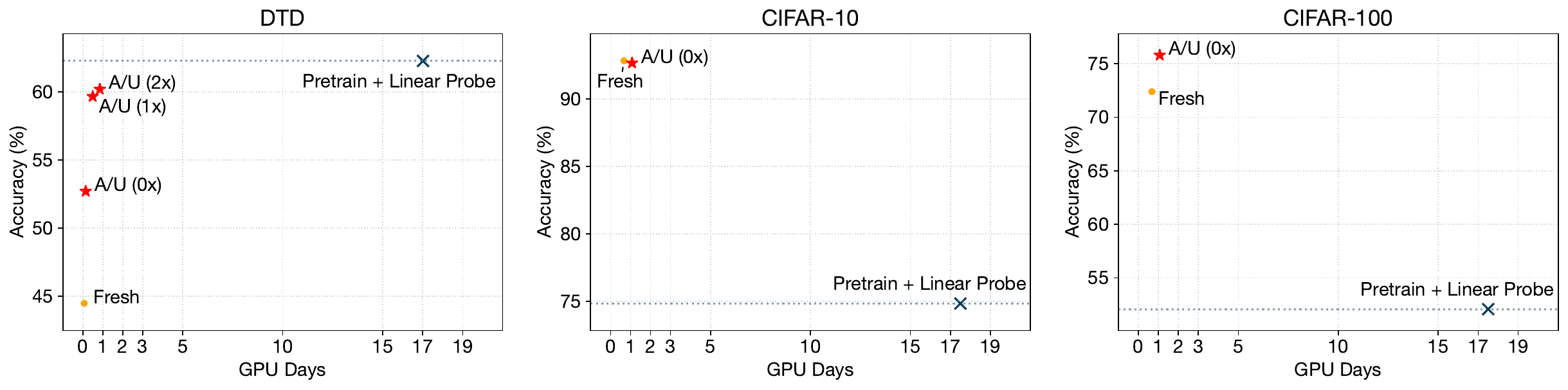} 
    \end{center}
    \caption{Cost/accuracy comparison of our method (A/U) to supervised training (Fresh) and pre-training then linear probing.
    The teacher was a ResNet50 and the student was a MobileNetV2. 
    All timing experiments were done on one NVIDIA P100 GPU. }
    \label{fig:costs}
\end{figure}


\section{Conclusion}
Mainstream attention continues to focus on large models. Fortunately, we show that small models can
leverage these advancements for their own benefit. By combining a pre-trained teacher, a novel knowledge transfer algorithm,
and augmenting the ``questions'' the student asks the teacher, any small model can significantly improve its
performance without weeks of pre-training; all that is required is a publicly available, pre-trained teacher, 
a contrastive learning algorithm, and a generative model. 
The demand for small models will always exist due to resource constraints or savings, and when combined with methods like NAS and pruning, small models have a bright future.


\subsubsection*{Acknowledgments}

This work utilizes resources supported by the National Science Foundation’s Major Research 
Instrumentation program, grant \#1725729, as well as the University of Illinois at Urbana-Champaign
\citep{kindratenko2020hal}.
SF was supported by the IBM-Illinois Discovery Accelerator Institute and the AMD Center of Excellence at the University of Illinois at Urbana-Champaign.

\bibliography{iclr2024_conference}

\begin{thebibliography}{80}
\providecommand{\natexlab}[1]{#1}
\providecommand{\url}[1]{\texttt{#1}}
\expandafter\ifx\csname urlstyle\endcsname\relax
  \providecommand{\doi}[1]{doi: #1}\else
  \providecommand{\doi}{doi: \begingroup \urlstyle{rm}\Url}\fi

\bibitem[Ahn et~al.(2019)Ahn, Hu, Damianou, Lawrence, and Dai]{ahn2019vid}
Sungsoo Ahn, Shell~Xu Hu, Andreas Damianou, Neil~D Lawrence, and Zhenwen Dai.
\newblock Variational information distillation for knowledge transfer.
\newblock In \emph{Proceedings of the IEEE/CVF conference on computer vision and pattern recognition}, pp.\  9163--9171, 2019.

\bibitem[Alayrac et~al.(2022)Alayrac, Donahue, Luc, Miech, Barr, Hasson, Lenc, Mensch, Millican, Reynolds, et~al.]{alayrac2022flamingo}
Jean-Baptiste Alayrac, Jeff Donahue, Pauline Luc, Antoine Miech, Iain Barr, Yana Hasson, Karel Lenc, Arthur Mensch, Katherine Millican, Malcolm Reynolds, et~al.
\newblock Flamingo: a visual language model for few-shot learning.
\newblock \emph{Advances in Neural Information Processing Systems}, 35:\penalty0 23716--23736, 2022.

\bibitem[Azizi et~al.(2023)Azizi, Kornblith, Saharia, Norouzi, and Fleet]{azizi2023synthetic}
Shekoofeh Azizi, Simon Kornblith, Chitwan Saharia, Mohammad Norouzi, and David~J Fleet.
\newblock Synthetic data from diffusion models improves imagenet classification.
\newblock \emph{arXiv preprint arXiv:2304.08466}, 2023.

\bibitem[Bansal \& Grover(2023)Bansal and Grover]{bansal2023leaving}
Hritik Bansal and Aditya Grover.
\newblock Leaving reality to imagination: Robust classification via generated datasets.
\newblock \emph{arXiv preprint arXiv:2302.02503}, 2023.

\bibitem[Bommasani et~al.(2021)Bommasani, Hudson, Adeli, Altman, Arora, von Arx, Bernstein, Bohg, Bosselut, Brunskill, et~al.]{bommasani2021foundations}
Rishi Bommasani, Drew~A Hudson, Ehsan Adeli, Russ Altman, Simran Arora, Sydney von Arx, Michael~S Bernstein, Jeannette Bohg, Antoine Bosselut, Emma Brunskill, et~al.
\newblock On the opportunities and risks of foundation models.
\newblock \emph{arXiv preprint arXiv:2108.07258}, 2021.

\bibitem[Brown et~al.(2020)Brown, Mann, Ryder, Subbiah, Kaplan, Dhariwal, Neelakantan, Shyam, Sastry, Askell, et~al.]{brown2020gpt3}
Tom Brown, Benjamin Mann, Nick Ryder, Melanie Subbiah, Jared~D Kaplan, Prafulla Dhariwal, Arvind Neelakantan, Pranav Shyam, Girish Sastry, Amanda Askell, et~al.
\newblock Language models are few-shot learners.
\newblock \emph{Advances in neural information processing systems}, 33:\penalty0 1877--1901, 2020.

\bibitem[Bucilua et~al.(2006)Bucilua, Caruana, and Niculescu-Mizil]{caruana2006compression}
Cristian Bucilua, Rich Caruana, and Alexandru Niculescu-Mizil.
\newblock Model compression.
\newblock In \emph{Proceedings of the 12th ACM SIGKDD International Conference on Knowledge Discovery and Data Mining}, KDD '06, pp.\  535–541, New York, NY, USA, 2006. Association for Computing Machinery.
\newblock ISBN 1595933395.
\newblock \doi{10.1145/1150402.1150464}.
\newblock URL \url{https://doi.org/10.1145/1150402.1150464}.

\bibitem[Chen et~al.(2021{\natexlab{a}})Chen, Mei, Zhang, Wang, Wang, Feng, and Chen]{chen2021semckd}
Defang Chen, Jian-Ping Mei, Yuan Zhang, Can Wang, Zhe Wang, Yan Feng, and Chun Chen.
\newblock Cross-layer distillation with semantic calibration.
\newblock In \emph{Proceedings of the AAAI Conference on Artificial Intelligence}, volume~35, pp.\  7028--7036, 2021{\natexlab{a}}.

\bibitem[Chen et~al.(2022)Chen, Mei, Zhang, Wang, Feng, and Chen]{chen2022simkd}
Defang Chen, Jian-Ping Mei, Hailin Zhang, Can Wang, Yan Feng, and Chun Chen.
\newblock Knowledge distillation with the reused teacher classifier.
\newblock In \emph{Proceedings of the IEEE/CVF Conference on Computer Vision and Pattern Recognition}, pp.\  11933--11942, 2022.

\bibitem[Chen et~al.(2021{\natexlab{b}})Chen, Liu, Zhao, and Jia]{chen2021reviewkd}
Pengguang Chen, Shu Liu, Hengshuang Zhao, and Jiaya Jia.
\newblock Distilling knowledge via knowledge review.
\newblock In \emph{Proceedings of the IEEE/CVF Conference on Computer Vision and Pattern Recognition}, pp.\  5008--5017, 2021{\natexlab{b}}.

\bibitem[Chen et~al.(2020{\natexlab{a}})Chen, Kornblith, Norouzi, and Hinton]{chen2020simclr}
Ting Chen, Simon Kornblith, Mohammad Norouzi, and Geoffrey Hinton.
\newblock A simple framework for contrastive learning of visual representations.
\newblock In \emph{International conference on machine learning}, pp.\  1597--1607. PMLR, 2020{\natexlab{a}}.

\bibitem[Chen \& He(2021)Chen and He]{chen2021simsiam}
Xinlei Chen and Kaiming He.
\newblock Exploring simple siamese representation learning.
\newblock In \emph{Proceedings of the IEEE/CVF conference on computer vision and pattern recognition}, pp.\  15750--15758, 2021.

\bibitem[Chen et~al.(2020{\natexlab{b}})Chen, Fan, Girshick, and He]{chen2020mocov2}
Xinlei Chen, Haoqi Fan, Ross Girshick, and Kaiming He.
\newblock Improved baselines with momentum contrastive learning.
\newblock \emph{arXiv preprint arXiv:2003.04297}, 2020{\natexlab{b}}.

\bibitem[Chopra et~al.(2005)Chopra, Hadsell, and LeCun]{contrastive_loss}
S.~Chopra, R.~Hadsell, and Y.~LeCun.
\newblock Learning a similarity metric discriminatively, with application to face verification.
\newblock In \emph{2005 IEEE Computer Society Conference on Computer Vision and Pattern Recognition (CVPR'05)}, volume~1, pp.\  539--546 vol. 1, 2005.
\newblock \doi{10.1109/CVPR.2005.202}.

\bibitem[Cimpoi et~al.(2014)Cimpoi, Maji, Kokkinos, Mohamed, , and Vedaldi]{cimpoi14dtd}
M.~Cimpoi, S.~Maji, I.~Kokkinos, S.~Mohamed, , and A.~Vedaldi.
\newblock Describing textures in the wild.
\newblock In \emph{Proceedings of the {IEEE} Conf. on Computer Vision and Pattern Recognition ({CVPR})}, 2014.

\bibitem[Deng et~al.(2009)Deng, Dong, Socher, Li, Li, and Fei-Fei]{imagenet}
Jia Deng, Wei Dong, Richard Socher, Li-Jia Li, Kai Li, and Li~Fei-Fei.
\newblock Imagenet: A large-scale hierarchical image database.
\newblock In \emph{2009 IEEE Conference on Computer Vision and Pattern Recognition}, pp.\  248--255, 2009.
\newblock \doi{10.1109/CVPR.2009.5206848}.

\bibitem[Dietterich(2000)]{Dietterich2000EnsembleMI}
Thomas~G. Dietterich.
\newblock Ensemble methods in machine learning.
\newblock In \emph{International Workshop on Multiple Classifier Systems}, 2000.

\bibitem[Dosovitskiy et~al.(2020)Dosovitskiy, Beyer, Kolesnikov, Weissenborn, Zhai, Unterthiner, Dehghani, Minderer, Heigold, Gelly, et~al.]{dosovitskiy2020vit}
Alexey Dosovitskiy, Lucas Beyer, Alexander Kolesnikov, Dirk Weissenborn, Xiaohua Zhai, Thomas Unterthiner, Mostafa Dehghani, Matthias Minderer, Georg Heigold, Sylvain Gelly, et~al.
\newblock An image is worth 16x16 words: Transformers for image recognition at scale.
\newblock \emph{arXiv preprint arXiv:2010.11929}, 2020.

\bibitem[Fang et~al.(2021)Fang, Wang, Wang, Zhang, Yang, and Liu]{fang2021seed}
Zhiyuan Fang, Jianfeng Wang, Lijuan Wang, Lei Zhang, Yezhou Yang, and Zicheng Liu.
\newblock {\{}SEED{\}}: Self-supervised distillation for visual representation.
\newblock In \emph{International Conference on Learning Representations}, 2021.
\newblock URL \url{https://openreview.net/forum?id=AHm3dbp7D1D}.

\bibitem[Frosst et~al.(2019)Frosst, Papernot, and Hinton]{frosst2019_softnn2}
Nicholas Frosst, Nicolas Papernot, and Geoffrey Hinton.
\newblock Analyzing and improving representations with the soft nearest neighbor loss.
\newblock In \emph{International conference on machine learning}, pp.\  2012--2020. PMLR, 2019.

\bibitem[Gao et~al.(2021)Gao, Zhuang, Lin, Cheng, Sun, Li, and Shen]{gao2021disco}
Yuting Gao, Jia-Xin Zhuang, Shaohui Lin, Hao Cheng, Xing Sun, Ke~Li, and Chunhua Shen.
\newblock Disco: Remedy self-supervised learning on lightweight models with distilled contrastive learning.
\newblock \emph{arXiv preprint arXiv:2104.09124}, 2021.

\bibitem[Grill et~al.(2020)Grill, Strub, Altch{\'e}, Tallec, Richemond, Buchatskaya, Doersch, Avila~Pires, Guo, Gheshlaghi~Azar, et~al.]{grill2020byol}
Jean-Bastien Grill, Florian Strub, Florent Altch{\'e}, Corentin Tallec, Pierre Richemond, Elena Buchatskaya, Carl Doersch, Bernardo Avila~Pires, Zhaohan Guo, Mohammad Gheshlaghi~Azar, et~al.
\newblock Bootstrap your own latent-a new approach to self-supervised learning.
\newblock \emph{Advances in neural information processing systems}, 33:\penalty0 21271--21284, 2020.

\bibitem[Gutmann \& Hyvärinen(2010)Gutmann and Hyvärinen]{pmlr-v9-gutmann10_nce}
Michael Gutmann and Aapo Hyvärinen.
\newblock Noise-contrastive estimation: A new estimation principle for unnormalized statistical models.
\newblock In Yee~Whye Teh and Mike Titterington (eds.), \emph{Proceedings of the Thirteenth International Conference on Artificial Intelligence and Statistics}, volume~9 of \emph{Proceedings of Machine Learning Research}, pp.\  297--304, Chia Laguna Resort, Sardinia, Italy, 13--15 May 2010. PMLR.
\newblock URL \url{https://proceedings.mlr.press/v9/gutmann10a.html}.

\bibitem[Han et~al.(2016)Han, Mao, and Dally]{han2016pruning}
Song Han, Huizi Mao, and William~J. Dally.
\newblock Deep compression: Compressing deep neural networks with pruning, trained quantization and huffman coding, 2016.

\bibitem[He et~al.(2016)He, Zhang, Ren, and Sun]{he2016resnet}
Kaiming He, Xiangyu Zhang, Shaoqing Ren, and Jian Sun.
\newblock Deep residual learning for image recognition.
\newblock In \emph{Proceedings of the IEEE conference on computer vision and pattern recognition}, pp.\  770--778, 2016.

\bibitem[He et~al.(2020)He, Fan, Wu, Xie, and Girshick]{he2020moco}
Kaiming He, Haoqi Fan, Yuxin Wu, Saining Xie, and Ross Girshick.
\newblock Momentum contrast for unsupervised visual representation learning.
\newblock In \emph{Proceedings of the IEEE/CVF conference on computer vision and pattern recognition}, pp.\  9729--9738, 2020.

\bibitem[He et~al.(2022)He, Sun, Yu, Xue, Zhang, Torr, Bai, and Qi]{he2022synthetic}
Ruifei He, Shuyang Sun, Xin Yu, Chuhui Xue, Wenqing Zhang, Philip Torr, Song Bai, and Xiaojuan Qi.
\newblock Is synthetic data from generative models ready for image recognition?
\newblock \emph{arXiv preprint arXiv:2210.07574}, 2022.

\bibitem[Heo et~al.(2019{\natexlab{a}})Heo, Kim, Yun, Park, Kwak, and Choi]{heo2019ofd}
Byeongho Heo, Jeesoo Kim, Sangdoo Yun, Hyojin Park, Nojun Kwak, and Jin~Young Choi.
\newblock A comprehensive overhaul of feature distillation.
\newblock In \emph{Proceedings of the IEEE/CVF International Conference on Computer Vision}, pp.\  1921--1930, 2019{\natexlab{a}}.

\bibitem[Heo et~al.(2019{\natexlab{b}})Heo, Lee, Yun, and Choi]{heo2019ab}
Byeongho Heo, Minsik Lee, Sangdoo Yun, and Jin~Young Choi.
\newblock Knowledge transfer via distillation of activation boundaries formed by hidden neurons.
\newblock In \emph{Proceedings of the AAAI Conference on Artificial Intelligence}, volume~33, pp.\  3779--3787, 2019{\natexlab{b}}.

\bibitem[Hinton et~al.(2015)Hinton, Vinyals, and Dean]{hinton2015distilling}
Geoffrey Hinton, Oriol Vinyals, and Jeff Dean.
\newblock Distilling the knowledge in a neural network, 2015.

\bibitem[Ho et~al.(2020)Ho, Jain, and Abbeel]{ho2020ddpm}
Jonathan Ho, Ajay Jain, and Pieter Abbeel.
\newblock Denoising diffusion probabilistic models.
\newblock \emph{Advances in neural information processing systems}, 33:\penalty0 6840--6851, 2020.

\bibitem[Huang \& Wang(2017)Huang and Wang]{huang2017nst}
Zehao Huang and Naiyan Wang.
\newblock Like what you like: Knowledge distill via neuron selectivity transfer.
\newblock \emph{arXiv preprint arXiv:1707.01219}, 2017.

\bibitem[Hyv{{\"a}}rinen(2005)]{JMLR:v6:hyvarinen05-scorematching}
Aapo Hyv{{\"a}}rinen.
\newblock Estimation of non-normalized statistical models by score matching.
\newblock \emph{Journal of Machine Learning Research}, 6\penalty0 (24):\penalty0 695--709, 2005.
\newblock URL \url{http://jmlr.org/papers/v6/hyvarinen05a.html}.

\bibitem[Jacob et~al.(2017)Jacob, Kligys, Chen, Zhu, Tang, Howard, Adam, and Kalenichenko]{jacob2017quantization}
Benoit Jacob, Skirmantas Kligys, Bo~Chen, Menglong Zhu, Matthew Tang, Andrew Howard, Hartwig Adam, and Dmitry Kalenichenko.
\newblock Quantization and training of neural networks for efficient integer-arithmetic-only inference, 2017.

\bibitem[Kim et~al.(2018)Kim, Park, and Kwak]{kim2018ft}
Jangho Kim, SeongUk Park, and Nojun Kwak.
\newblock Paraphrasing complex network: Network compression via factor transfer.
\newblock \emph{Advances in neural information processing systems}, 31, 2018.

\bibitem[Kindratenko et~al.(2020)Kindratenko, Mu, Zhan, Maloney, Hashemi, Rabe, Xu, Campbell, Peng, and Gropp]{kindratenko2020hal}
Volodymyr Kindratenko, Dawei Mu, Yan Zhan, John Maloney, Sayed~Hadi Hashemi, Benjamin Rabe, Ke~Xu, Roy Campbell, Jian Peng, and William Gropp.
\newblock Hal: Computer system for scalable deep learning.
\newblock In \emph{Practice and experience in advanced research computing}, pp.\  41--48. 2020.

\bibitem[Krizhevsky et~al.(2009)Krizhevsky, Nair, and Hinton]{krizhevsky2009cifar}
Alex Krizhevsky, Vinod Nair, and Geoffrey Hinton.
\newblock Cifar-10 and cifar-100 datasets.
\newblock \emph{URl: https://www. cs. toronto. edu/kriz/cifar. html}, 6\penalty0 (1):\penalty0 1, 2009.

\bibitem[Li et~al.(2003)Li, Andreetto, Ranzato, and Perona]{li2003caltech}
FF~Li, M~Andreetto, MA~Ranzato, and P~Perona.
\newblock Caltech 101.
\newblock \emph{Computational Vision Group, California Institute of Technology}, 7, 2003.

\bibitem[Li \& Hoiem(2017)Li and Hoiem]{li2017lwf}
Zhizhong Li and Derek Hoiem.
\newblock Learning without forgetting.
\newblock \emph{IEEE transactions on pattern analysis and machine intelligence}, 40\penalty0 (12):\penalty0 2935--2947, 2017.

\bibitem[Liu et~al.(2019)Liu, Simonyan, and Yang]{liu2019darts}
Hanxiao Liu, Karen Simonyan, and Yiming Yang.
\newblock Darts: Differentiable architecture search, 2019.

\bibitem[Liu et~al.(2021)Liu, Nie, Yin, Wang, Gao, and Jin]{liu2021sskd}
Weidong Liu, Shibo Nie, Junhui Yin, Rui Wang, Donghui Gao, and Ling Jin.
\newblock Sskd: Self-supervised knowledge distillation for cross domain adaptive person re-identification.
\newblock In \emph{2021 7th IEEE International Conference on Network Intelligence and Digital Content (IC-NIDC)}, pp.\  81--85. IEEE, 2021.

\bibitem[Loshchilov \& Hutter(2019)Loshchilov and Hutter]{loshchilov2018adamw}
Ilya Loshchilov and Frank Hutter.
\newblock Decoupled weight decay regularization.
\newblock In \emph{International Conference on Learning Representations}, 2019.
\newblock URL \url{https://openreview.net/forum?id=Bkg6RiCqY7}.

\bibitem[Miles et~al.(2021)Miles, Rodriguez, and Mikolajczyk]{miles2021itrd}
Roy Miles, Adrian~Lopez Rodriguez, and Krystian Mikolajczyk.
\newblock Information theoretic representation distillation.
\newblock \emph{arXiv preprint arXiv:2112.00459}, 2021.

\bibitem[Oord et~al.(2018)Oord, Li, and Vinyals]{oord2018representation_infonce}
Aaron van~den Oord, Yazhe Li, and Oriol Vinyals.
\newblock Representation learning with contrastive predictive coding.
\newblock \emph{arXiv preprint arXiv:1807.03748}, 2018.

\bibitem[Park et~al.(2019)Park, Kim, Lu, and Cho]{park2019rkd}
Wonpyo Park, Dongju Kim, Yan Lu, and Minsu Cho.
\newblock Relational knowledge distillation.
\newblock In \emph{Proceedings of the IEEE/CVF conference on computer vision and pattern recognition}, pp.\  3967--3976, 2019.

\bibitem[Passalis \& Tefas(2018)Passalis and Tefas]{passalis2018pkt}
Nikolaos Passalis and Anastasios Tefas.
\newblock Learning deep representations with probabilistic knowledge transfer.
\newblock In \emph{Proceedings of the European Conference on Computer Vision (ECCV)}, pp.\  268--284, 2018.

\bibitem[Peng et~al.(2019)Peng, Jin, Liu, Li, Wu, Liu, Zhou, and Zhang]{peng2019cc}
Baoyun Peng, Xiao Jin, Jiaheng Liu, Dongsheng Li, Yichao Wu, Yu~Liu, Shunfeng Zhou, and Zhaoning Zhang.
\newblock Correlation congruence for knowledge distillation.
\newblock In \emph{Proceedings of the IEEE/CVF International Conference on Computer Vision}, pp.\  5007--5016, 2019.

\bibitem[Quattoni \& Torralba(2009)Quattoni and Torralba]{quattoni2009mit_indoor}
Ariadna Quattoni and Antonio Torralba.
\newblock Recognizing indoor scenes.
\newblock In \emph{2009 IEEE conference on computer vision and pattern recognition}, pp.\  413--420. IEEE, 2009.

\bibitem[Radford et~al.(2019)Radford, Wu, Child, Luan, Amodei, and Sutskever]{radford2019gpt2}
Alec Radford, Jeff Wu, Rewon Child, David Luan, Dario Amodei, and Ilya Sutskever.
\newblock Language models are unsupervised multitask learners.
\newblock 2019.

\bibitem[Radford et~al.(2021)Radford, Kim, Hallacy, Ramesh, Goh, Agarwal, Sastry, Askell, Mishkin, Clark, et~al.]{radford2021clip}
Alec Radford, Jong~Wook Kim, Chris Hallacy, Aditya Ramesh, Gabriel Goh, Sandhini Agarwal, Girish Sastry, Amanda Askell, Pamela Mishkin, Jack Clark, et~al.
\newblock Learning transferable visual models from natural language supervision.
\newblock In \emph{International conference on machine learning}, pp.\  8748--8763. PMLR, 2021.

\bibitem[Rombach et~al.(2022)Rombach, Blattmann, Lorenz, Esser, and Ommer]{rombach2022ldm}
Robin Rombach, Andreas Blattmann, Dominik Lorenz, Patrick Esser, and Bj{\"o}rn Ommer.
\newblock High-resolution image synthesis with latent diffusion models.
\newblock In \emph{Proceedings of the IEEE/CVF Conference on Computer Vision and Pattern Recognition}, pp.\  10684--10695, 2022.

\bibitem[Romero et~al.(2014)Romero, Ballas, Kahou, Chassang, Gatta, and Bengio]{romero2014fitnets}
Adriana Romero, Nicolas Ballas, Samira~Ebrahimi Kahou, Antoine Chassang, Carlo Gatta, and Yoshua Bengio.
\newblock Fitnets: Hints for thin deep nets.
\newblock \emph{arXiv preprint arXiv:1412.6550}, 2014.

\bibitem[Russakovsky et~al.(2015)Russakovsky, Deng, Su, Krause, Satheesh, Ma, Huang, Karpathy, Khosla, Bernstein, et~al.]{russakovsky2015ilsvrc}
Olga Russakovsky, Jia Deng, Hao Su, Jonathan Krause, Sanjeev Satheesh, Sean Ma, Zhiheng Huang, Andrej Karpathy, Aditya Khosla, Michael Bernstein, et~al.
\newblock Imagenet large scale visual recognition challenge.
\newblock \emph{International journal of computer vision}, 115:\penalty0 211--252, 2015.

\bibitem[Salakhutdinov \& Hinton(2007)Salakhutdinov and Hinton]{pmlr-v2-salakhutdinov07_softnn}
Ruslan Salakhutdinov and Geoff Hinton.
\newblock Learning a nonlinear embedding by preserving class neighbourhood structure.
\newblock In Marina Meila and Xiaotong Shen (eds.), \emph{Proceedings of the Eleventh International Conference on Artificial Intelligence and Statistics}, volume~2 of \emph{Proceedings of Machine Learning Research}, pp.\  412--419, San Juan, Puerto Rico, 21--24 Mar 2007. PMLR.
\newblock URL \url{https://proceedings.mlr.press/v2/salakhutdinov07a.html}.

\bibitem[Salimans \& Ho(2022)Salimans and Ho]{salimans2022progressive}
Tim Salimans and Jonathan Ho.
\newblock Progressive distillation for fast sampling of diffusion models.
\newblock \emph{arXiv preprint arXiv:2202.00512}, 2022.

\bibitem[Sandler et~al.(2018)Sandler, Howard, Zhu, Zhmoginov, and Chen]{sandler2018mobilenetv2}
Mark Sandler, Andrew Howard, Menglong Zhu, Andrey Zhmoginov, and Liang-Chieh Chen.
\newblock Mobilenetv2: Inverted residuals and linear bottlenecks.
\newblock In \emph{Proceedings of the IEEE conference on computer vision and pattern recognition}, pp.\  4510--4520, 2018.

\bibitem[Sariyildiz et~al.(2022)Sariyildiz, Alahari, Larlus, and Kalantidis]{sariyildiz2022fake}
Mert~Bulent Sariyildiz, Karteek Alahari, Diane Larlus, and Yannis Kalantidis.
\newblock Fake it till you make it: Learning (s) from a synthetic imagenet clone.
\newblock \emph{arXiv preprint arXiv:2212.08420}, 2022.

\bibitem[Schroff et~al.(2015)Schroff, Kalenichenko, and Philbin]{schroff2015triplet}
Florian Schroff, Dmitry Kalenichenko, and James Philbin.
\newblock Facenet: A unified embedding for face recognition and clustering.
\newblock In \emph{Proceedings of the IEEE conference on computer vision and pattern recognition}, pp.\  815--823, 2015.

\bibitem[Schuhmann et~al.(2022)Schuhmann, Beaumont, Vencu, Gordon, Wightman, Cherti, Coombes, Katta, Mullis, Wortsman, et~al.]{schuhmann2022laion}
Christoph Schuhmann, Romain Beaumont, Richard Vencu, Cade Gordon, Ross Wightman, Mehdi Cherti, Theo Coombes, Aarush Katta, Clayton Mullis, Mitchell Wortsman, et~al.
\newblock Laion-5b: An open large-scale dataset for training next generation image-text models.
\newblock \emph{Advances in Neural Information Processing Systems}, 35:\penalty0 25278--25294, 2022.

\bibitem[Sohl-Dickstein et~al.(2015)Sohl-Dickstein, Weiss, Maheswaranathan, and Ganguli]{sohl2015deep}
Jascha Sohl-Dickstein, Eric Weiss, Niru Maheswaranathan, and Surya Ganguli.
\newblock Deep unsupervised learning using nonequilibrium thermodynamics.
\newblock In \emph{International conference on machine learning}, pp.\  2256--2265. PMLR, 2015.

\bibitem[Sohn(2016)]{NIPS2016_6b180037_npairs}
Kihyuk Sohn.
\newblock Improved deep metric learning with multi-class n-pair loss objective.
\newblock In D.~Lee, M.~Sugiyama, U.~Luxburg, I.~Guyon, and R.~Garnett (eds.), \emph{Advances in Neural Information Processing Systems}, volume~29. Curran Associates, Inc., 2016.
\newblock URL \url{https://proceedings.neurips.cc/paper_files/paper/2016/file/6b180037abbebea991d8b1232f8a8ca9-Paper.pdf}.

\bibitem[Song \& Ermon(2019)Song and Ermon]{song2019ncsn}
Yang Song and Stefano Ermon.
\newblock Generative modeling by estimating gradients of the data distribution.
\newblock \emph{Advances in neural information processing systems}, 32, 2019.

\bibitem[Song et~al.(2020)Song, Sohl-Dickstein, Kingma, Kumar, Ermon, and Poole]{song2020score}
Yang Song, Jascha Sohl-Dickstein, Diederik~P Kingma, Abhishek Kumar, Stefano Ermon, and Ben Poole.
\newblock Score-based generative modeling through stochastic differential equations.
\newblock \emph{arXiv preprint arXiv:2011.13456}, 2020.

\bibitem[Song et~al.(2023)Song, Dhariwal, Chen, and Sutskever]{song2023consistency}
Yang Song, Prafulla Dhariwal, Mark Chen, and Ilya Sutskever.
\newblock Consistency models.
\newblock 2023.

\bibitem[Srinivas \& Fleuret(2018)Srinivas and Fleuret]{srinivas2018jacobian}
Suraj Srinivas and Fran{\c{c}}ois Fleuret.
\newblock Knowledge transfer with jacobian matching.
\newblock In \emph{International Conference on Machine Learning}, pp.\  4723--4731. PMLR, 2018.

\bibitem[Tian et~al.(2020)Tian, Krishnan, and Isola]{tian2019crd}
Yonglong Tian, Dilip Krishnan, and Phillip Isola.
\newblock Contrastive representation distillation.
\newblock In \emph{International Conference on Learning Representations}, 2020.

\bibitem[Tung \& Mori(2019)Tung and Mori]{tung2019sp}
Frederick Tung and Greg Mori.
\newblock Similarity-preserving knowledge distillation.
\newblock In \emph{Proceedings of the IEEE/CVF international conference on computer vision}, pp.\  1365--1374, 2019.

\bibitem[Vincent(2011)]{denoising_scorematching}
Pascal Vincent.
\newblock {A Connection Between Score Matching and Denoising Autoencoders}.
\newblock \emph{Neural Computation}, 23\penalty0 (7):\penalty0 1661--1674, 07 2011.
\newblock ISSN 0899-7667.
\newblock \doi{10.1162/NECO_a_00142}.
\newblock URL \url{https://doi.org/10.1162/NECO\_a\_00142}.

\bibitem[Wah et~al.(2011)Wah, Branson, Welinder, Perona, and Belongie]{Wah2011cub_2011}
Catherine Wah, Steve Branson, Peter Welinder, Pietro Perona, and Serge~J. Belongie.
\newblock The caltech-ucsd birds-200-2011 dataset.
\newblock 2011.

\bibitem[Wang \& Isola(2020)Wang and Isola]{wang2020align_uniform}
Tongzhou Wang and Phillip Isola.
\newblock Understanding contrastive representation learning through alignment and uniformity on the hypersphere.
\newblock In \emph{International Conference on Machine Learning}, pp.\  9929--9939. PMLR, 2020.

\bibitem[Wu et~al.(2018)Wu, Xiong, Yu, and Lin]{wu2018unsupervised}
Zhirong Wu, Yuanjun Xiong, Stella Yu, and Dahua Lin.
\newblock Unsupervised feature learning via non-parametric instance-level discrimination, 2018.

\bibitem[Xu et~al.(2021)Xu, Fang, Zhang, Xie, Wang, Dai, Xiong, and Tian]{xu2021bingo}
Haohang Xu, Jiemin Fang, Xiaopeng Zhang, Lingxi Xie, Xinggang Wang, Wenrui Dai, Hongkai Xiong, and Qi~Tian.
\newblock Bag of instances aggregation boosts self-supervised distillation.
\newblock \emph{arXiv preprint arXiv:2107.01691}, 2021.

\bibitem[Yang et~al.(2021)Yang, Martinez, Bulat, and Tzimiropoulos]{yang2021srrl}
Jing Yang, Brais Martinez, Adrian Bulat, and Georgios Tzimiropoulos.
\newblock Knowledge distillation via softmax regression representation learning.
\newblock In \emph{ICLR2021}, 2021.

\bibitem[Yim et~al.(2017)Yim, Joo, Bae, and Kim]{fsp}
Junho Yim, Donggyu Joo, Jihoon Bae, and Junmo Kim.
\newblock A gift from knowledge distillation: Fast optimization, network minimization and transfer learning.
\newblock In \emph{2017 IEEE Conference on Computer Vision and Pattern Recognition (CVPR)}, pp.\  7130--7138, 2017.
\newblock \doi{10.1109/CVPR.2017.754}.

\bibitem[Yuan et~al.(2021)Yuan, Chen, Chen, Codella, Dai, Gao, Hu, Huang, Li, Li, et~al.]{yuan2021florence}
Lu~Yuan, Dongdong Chen, Yi-Ling Chen, Noel Codella, Xiyang Dai, Jianfeng Gao, Houdong Hu, Xuedong Huang, Boxin Li, Chunyuan Li, et~al.
\newblock Florence: A new foundation model for computer vision.
\newblock \emph{arXiv preprint arXiv:2111.11432}, 2021.

\bibitem[Zagoruyko \& Komodakis(2016)Zagoruyko and Komodakis]{zagoruyko2016at}
Sergey Zagoruyko and Nikos Komodakis.
\newblock Paying more attention to attention: Improving the performance of convolutional neural networks via attention transfer.
\newblock \emph{arXiv preprint arXiv:1612.03928}, 2016.

\bibitem[Zbontar et~al.(2021)Zbontar, Jing, Misra, LeCun, and Deny]{zbontar2021barlow}
Jure Zbontar, Li~Jing, Ishan Misra, Yann LeCun, and St{\'e}phane Deny.
\newblock Barlow twins: Self-supervised learning via redundancy reduction.
\newblock In \emph{International Conference on Machine Learning}, pp.\  12310--12320. PMLR, 2021.

\bibitem[Zhao et~al.(2022)Zhao, Cui, Song, Qiu, and Liang]{zhao2022dkd}
Borui Zhao, Quan Cui, Renjie Song, Yiyu Qiu, and Jiajun Liang.
\newblock Decoupled knowledge distillation.
\newblock In \emph{Proceedings of the IEEE/CVF Conference on computer vision and pattern recognition}, pp.\  11953--11962, 2022.

\bibitem[Zhou et~al.(2023)Zhou, Sahak, and Ba]{zhou2023training}
Yongchao Zhou, Hshmat Sahak, and Jimmy Ba.
\newblock Training on thin air: Improve image classification with generated data.
\newblock 2023.

\bibitem[Zoph \& Le(2017)Zoph and Le]{zoph2017nas}
Barret Zoph and Quoc~V. Le.
\newblock Neural architecture search with reinforcement learning, 2017.

\end{thebibliography}
\bibliographystyle{iclr2024_conference}

\appendix
\section{Appendix}

\subsection{Additional Results and Model Pairings}

In our results, we use several abbreviations for brevity. \textbf{FR} stands for a model that was initialized randomly and trained
end-to-end on the task. \textbf{LP} refers to a model that was pre-trained on ImageNet \citep{imagenet,russakovsky2015ilsvrc}
and linearly probed for the task. \textbf{T} and \textbf{S} refer to the teacher and student models, respectively. 
Note that all non-A/U trainings (i.e. finetuning the teacher, vanilla training the student), were done with just the dataset; no synthetic samples were involved.

\subsubsection{Self-Supervised Transfer Learning}
\label{sec:self-supervised-transfer}

In the self-supervised learning literature, pre-training is significantly less effective as model size decreases. 
SEED \citep{fang2021seed}, DisCo \citep{gao2021disco}, and BINGO \citep{xu2021bingo}
attempt to mitigate this by augmenting training with distillation from a pre-trained teacher backbone.
They test performance by training/distilling on ImageNet first, then adapting to tasks. For us, we sidestep ImageNet and
run our distillation directly on the downstream task.
We compare our strategy to theirs in Table \ref{table:self-supervised}.
We use the same teacher model, a pre-trained MoCoV2 ResNet50, and student, a ResNet18. 
The baseline is a ResNet18 pre-trained on ImageNet via MoCoV2 and linearly probed for a task. 
We omit the teacher's head to match their teacher-student setup ($\lambda_3 = 0$), so the knowledge transfer is
purely feature-based.
Our method surpasses prior
work by a large margin, lending support to the idea that small models may not need their own strong backbone, 
especially in the self-supervised domain where model size matters more.

\begin{table}[H]
    \caption{Comparing to self-supervised distillation.}
    \label{table:self-supervised}
    \begin{center}
    \begin{tabular}{c c c}
    Method  & CIFAR-10 & CIFAR-100 \bigstrut[b] \\
    \hline
    Baseline & 77.9 & 48.1 \bigstrut[t] \\
    SEED \citep{fang2021seed} & 82.3 & 56.8 \\
    DisCo \citep{gao2021disco} & 85.3 & 63.3 \\
    BINGO \citep{xu2021bingo} & 86.8 & 66.5 \bigstrut[b] \\
    \hline
    A/U (Ours) & \textbf{94.33}	& \textbf{73.83} \bigstrut[t]
    \end{tabular}
    \end{center}
\end{table}

\subsubsection{Accuracy Comparison}
\label{sec:acc_comparison}

We provide a tabular visualization of how our method compares to pre-training in Table \ref{table:gen_results}. On 4 out of 6 datasets, we are able to surpass the pre-training performance, while on the remaining two, which are considered quite data-limited, we come within $2\%$ when using the added synthetic data.

\begin{table}[H]
    \caption{Our method (A/U) can achieve performance close to, and often surpass, pre-training.}
    \label{table:gen_results}
    \begin{center}
    \begin{tabular}{c|cccccc}
        Method & CIFAR-10 & CIFAR-100 & MIT-67 & CUB-2011 & Caltech-101 & DTD  \bigstrut[b]\\
        \hline
        A/U ($0 \times$) & \textbf{92.66} & \textbf{75.8} & 67.09 & \textbf{67.45} & 87.74 & 52.71 \bigstrut[t] \\
        A/U ($1 \times$) & - & - & 71.57 & - & 89.94 & 59.68 \\
        A/U ($2 \times$) & - & - & \textbf{72.54} & - & 89.83 & 60.21 \bigstrut[b] \\
        \hline
        S-LP & 74.84 & 52.08 & 68.13 & 65.14 & \textbf{91.02} & \textbf{62.29} \bigstrut[t]
    \end{tabular} 
    \end{center}
\end{table}

\subsubsection{Comparison to Other Methods on Various Model Pairings}
\label{sec:additional_pairs}
For a fair comparison, we also test using a pre-trained
teacher with prior state-of-the-art KD algorithms
that did not consider our setting. Notably, we choose ones that have demonstrated
high efficacy as well as being architecture-agnostic; they do not utilize any intermediate feature information. The three works we compare to 
are: 

\begin{enumerate}
    \item KD \citep{hinton2015distilling}: The original KD loss that compares logits.
    \item CRD++ \citep{tian2019crd}: A contrastive-based distillation method. 
    We also include $\mathcal{L}_{\text{KD}}$. As their work pre-dated many
    modern advancements in contrastive learning, we compare our method to a modernized version which substitutes the 
    linear projection $g(\cdot)$ with the same 2-layer MLP we use in our A/U loss.
    \item SRRL \citep{yang2021srrl}: A method that passes the student features through the teacher classifier. 
        We do not use $\mathcal{L}_{\text{KD}}$ here
        as the original authors did not. However, they only considered convolutional networks and utilized a small convnet to project
        the student features to the teacher's dimension. We substitute this with an MLP to make it general to the underlying
        model architecture (Figure \ref{fig:srrl_connector}).
\end{enumerate}

The implementations of these methods were taken from their open-source repositories, which we thank the authors for making available, and
adapted appropriately. These results can be found in Tables \ref{table:r50_to_mbv2}, \ref{table:vit_to_mbv2}, \ref{table:r50_to_r18}, and \ref{table:vit_to_r18}.
In all setups, our algorithm (A/U) leads to high performance gains that are competitive with, and sometimes surpass, prior work. 
On one hand, these results demonstrate the superiority of our A/U loss in certain setups; on the other hand, many existing KD algorithms
can be substituted in $\mathcal{L}$ for our overall paradigm and still offer competitive boosts to small model performance. 
We also emphasize that, while CRD++ is also a contrastive method, our A/U method is computationally cheaper due to the lack of a negative sample bank, while often outperforming it.

\begin{table}[H]
    \caption{Accuracies (\%) of MobileNetV2 when assisted by a pre-trained ResNet50.}
    \label{table:r50_to_mbv2}
    \begin{center}
    \begin{tabular}{c|c c c c c c}
    Method & CIFAR-10 & CIFAR-100 & MIT-67 & CUB-2011 & Caltech-101 & DTD \bigstrut[b] \\
    \hline
    T-LP & 77.3 & 53.65 & 72.39 & 63.26 & 92.94 & 65.96 \bigstrut[t] \\
    S-FR & 92.83 & 72.39 & 57.84 & 58.23 & 81.13 & 44.47 \bigstrut[b] \\
    \hline 
    KD & 92.92 & 75.97 & 66.04 & \textbf{67.48} & 86.61 & 52.02 \bigstrut[t] \\
    SRRL & 92.85 & 73.24 & 65.67 & 62.81 & 85.31 & 49.79 \\
    CRD++ & \textbf{93.45} & \textbf{76.37} & 65.75 & 67.21 & 86.89 & 52.45 \bigstrut[b] \\
    \hline 
    \textbf{A/U (Ours)} & 92.66 & 75.8 & \textbf{67.09} & 67.45 & \textbf{87.74} & \textbf{52.71} \bigstrut[t] \\
    \hline
    S-LP & 74.84 & 52.08 & 68.13 & 65.14 & 91.02 & 62.29 \bigstrut[t]
    \end{tabular}
    \end{center}
\end{table}

\begin{table}[H]
    \caption{Accuracies (\%) of MobileNetV2 when assisted by a pre-trained ViT-B-16.}
    \label{table:vit_to_mbv2}
    \begin{center}
    \begin{tabular}{c|c c c c c c}
    Method & CIFAR-10 & CIFAR-100 & MIT-67 & CUB-2011 & Caltech-101 & DTD \bigstrut[b] \\
    \hline 
    T-LP & 95.14 & 80.56 & 81.34 & 79.06 & 94.80 & 70.21 \bigstrut[t] \\
    S-FR & 92.83 & 72.39 & 57.84 & 58.23 & 81.13 & 44.47 \bigstrut[b] \\
    \hline 
    KD & 94.33 & 78.9 & \textbf{66.34} & \textbf{71.16} & 84.92 & \textbf{49.79} \bigstrut[t] \\
    SRRL & 93.99 & 78.66 & 63.51 & 67.66 & \textbf{85.31} & 49.73 \\
    CRD++ & 93.93 & 78.57 & 65.37 & 70.45 & 82.49 & 48.4 \bigstrut[b] \\
    \hline 
    \textbf{A/U (Ours)} & \textbf{94.55} & \textbf{79.71} & 65.97 & 70.87 & 84.8 & 49.47 \bigstrut[t] \\
    \hline
    S-LP & 74.84 & 52.08 & 68.13 & 65.14 & 91.02 & 62.29 \bigstrut[t]
    \end{tabular}
    \end{center}
\end{table}

\begin{table}[H]
    \caption{Accuracies (\%) of ResNet18 when assisted by a pre-trained ResNet50.}
    \label{table:r50_to_r18}
    \begin{center}
    \begin{tabular}{c|c c c c c c}
    Method & CIFAR-10 & CIFAR-100 & MIT-67 & CUB-2011 & Caltech-101 & DTD \bigstrut[b] \\
    \hline
    T-LP & 77.3 & 53.65 & 72.39 & 63.26 & 92.94 & 65.96 \bigstrut[t] \\
    S-FR & 93.22 & 71.44 & 49.85 & 49.17 & 77.40 & 34.47 \bigstrut[b] \\
    \hline 
    KD & 92.48 & \textbf{75.94} & 65.22 & \textbf{64.77} & 85.14 & 46.44 \bigstrut[t] \\
    SRRL & \textbf{92.93} & 72.06 & 64.63 & 62.74 & \textbf{86.27} & 47.82 \\
    CRD++ & 92.87 & 75.72 & 65.75 & 64.19 & 85.25 & \textbf{48.78} \bigstrut[b] \\
    \hline 
    \textbf{A/U (Ours)} & 92.79 & 75.11 & \textbf{66.72} & 63.88 & 85.54 & 46.01 \bigstrut[c] \\
    \hline 
    S-LP & 78.38 & 55.44 & 66.27 & 62.75 & 90.23 & 62.07 \bigstrut[t]
    \end{tabular}
    \end{center}
\end{table}

\begin{table}[H]
    \caption{Accuracies (\%) of ResNet18 when assisted by a pre-trained ViT-B-16.}
    \label{table:vit_to_r18}
    \begin{center}
    \begin{tabular}{c|c c c c c c}
    Method & CIFAR-10 & CIFAR-100 & MIT-67 & CUB-2011 & Caltech-101 & DTD \bigstrut[b] \\
    \hline
    T-LP & 95.14 & 80.56 & 81.34 & 79.06 & 94.80 & 70.21 \bigstrut[t] \\
    S-FR & 93.22 & 71.44 & 49.85 & 49.17 & 77.40 & 34.47 \bigstrut[b] \\
    \hline 
    KD & 94.54 & 78.95 & \textbf{65.82} & \textbf{68.81} & 83.28 & \textbf{44.15} \bigstrut[t] \\
    SRRL & 94.03 & 77.46 & 61.87 & 62.94 & 81.58 & 37.66 \\
    CRD++ & 93.94 & \textbf{79.06} & 63.28 & 67.35 & 79.49 & 42.61 \bigstrut[b] \\
    \hline 
    \textbf{A/U (Ours)} & \textbf{94.98} & 78.89 & 65.75 & 68.48 & \textbf{83.39} & 40.69 \bigstrut[c] \\
    \hline 
    S-LP & 78.38 & 55.44 & 66.27 & 62.75 & 90.23 & 62.07 \bigstrut[t]
    \end{tabular}
    \end{center}
\end{table}

\subsection{Details of Our Loss}

Our contrastive-based loss (Equation \ref{eq:kd_contrastive}) can be interpreted as translating the InfoNCE loss of \citet{oord2018representation_infonce} to the
distillation case:

\begin{equation}
    \label{eq:infonce}
    \tag{InfoNCE}
    \mathbb{E}_{(x, y) \sim p_{\text{pos}}, \{x_i^-\}_{i=0}^M \sim p_{\text{data}}} \left[ - \log \frac{e^{f(x)^\top f(y) / \tau}}
            {e^{f(x)^\top f(y) / \tau} + \sum_i e^{f(x_i^-)^\top f(y) / \tau}} \right]
\end{equation}

We also employ the original method of \citet{hinton2015distilling}, which compares the student and teacher logits 
(pre-softmax vectors) via the cross-entropy:

\begin{equation}
    \label{eq:kd}
   \mathcal{L}_{KD}(z_s, z_t) = \tau^2\mathcal{H}(\sigma(z_t / \tau), \sigma(z_s / \tau))
\end{equation}

where $\sigma$ is the softmax function and $\tau$ is a temperature hyperparameter that controls the smoothness of the softmax; a higher
$\tau$ leads to a more uniform distribution.

Our Alignment/Uniformity setup is a re-formulation of \citet{wang2020align_uniform} for standard contrastive learning:

\begin{align}
\mathcal{L}_{\text{align}}(f; \alpha) &:= \mathbb{E}_{(x,y)\sim p_{\text{pos}}}[||f(x) - f(y)||_2^\alpha], \alpha > 0 \\
\mathcal{L}_{\text{uniform}}(f;t) &:= \log \mathbb{E}_{(x, y) \sim p_{\text{data}}} [e^{-t ||f(x) - f(y)||_2^2}], t > 0
\end{align}

\newcommand{\var}[1]{{\ttfamily#1}}
\begin{figure}[H]
    \centering
    \begin{subfigure}{0.49\textwidth} 
        \centering
\begin{lstlisting}[language=Python,frame=single,basicstyle=\ttfamily\small]
# x: batch of samples
# y: batch of other positive half

import torch
def align_loss(x, y, alpha=2):
    dists = (x - y).norm(p=2, dim=1) 
    return dists.pow(alpha).mean()

def unif_loss(x, t=2):
    pair = -t*torch.pdist(x, p=2)**2
    return pair.exp().mean().log()
\end{lstlisting}
            \begin{algorithmic}[1]
            \Require $f_S, f_T, g_S, g_T$: models and projectors
            \Require $w_a, w_u$: loss weights
                \State Sample batch $x$
                \State $z_S \gets g_S \circ f_S(x), z_T \gets g_T \circ f_T(x)$
                \State $l_a = $ \var{align\_loss}($z_S, z_T$)
                \State $l_u = \frac{1}{2}($\var{unif\_loss}$(z_S) + $\var{unif\_loss}$(z_T))$
                \State $\mathcal{L}_{\text{Embed}} = w_a l_a + w_u l_u$
            \end{algorithmic}
        \caption{Pseudocode}
        \label{fig:kd_code}
    \end{subfigure}
    \begin{subfigure}{0.49\textwidth}
        \centering
        \includegraphics{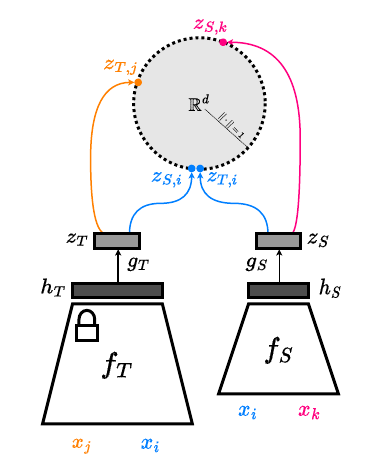}
        \caption{Flow chart}
        \label{fig:kd_flow}
    \end{subfigure}
    \caption{Our Alignment/Uniformity (A/U) based contrastive loss.}
\end{figure}

\subsection{Synthetic Data Generation Details}
\label{sec:syn_details}

\begin{figure}[H]
 \centering
 \subfloat[``the inside of a dental office'']{%
      \includegraphics[width=0.25\textwidth]{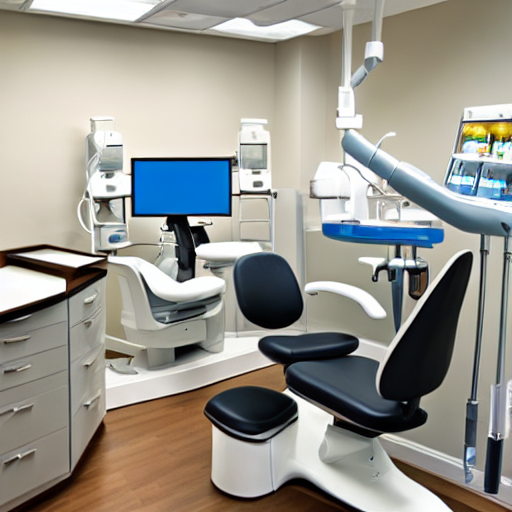}}
      \label{fig:image-a}
 \qquad
 \subfloat[``knitted texture'']{%
      \includegraphics[width=0.25\textwidth]{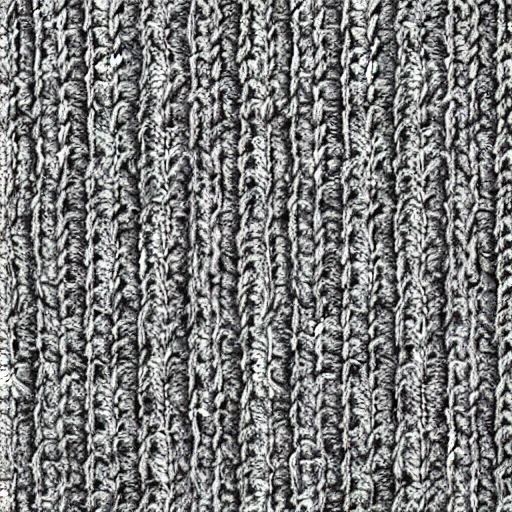}}
      \label{fig:image-b}
 \qquad
 \subfloat[``a picture of a tiger'']{%
      \includegraphics[width=0.25\textwidth]{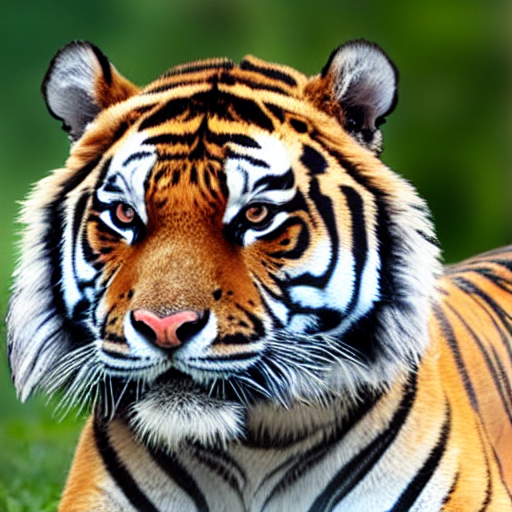}}
      \label{fig:image-c}
 \caption{Examples of synthetic samples generated for each dataset, (a) MIT-67, (b) DTD, (c) Caltech-101, with the respective prompt.}
 \label{fig:syn_samples}
\end{figure}

We find that 3 datasets require synthetic data to get close to pre-train levels of performance: MIT-67 \citep{quattoni2009mit_indoor}, 
Caltech-101 \citep{li2003caltech}, and DTD \citep{cimpoi14dtd}, all of which are considered data-limited. 
The diffusion model was set to take 50 inference steps per batch of images, amounting to around 12.3 seconds per image on 1 NVIDIA P100 GPU.
Our choice of prompts and timing numbers can be found in Table \ref{table:prompts}. Samples of generated images can be found in \ref{fig:syn_samples}.

\begin{table}[H]
    \vspace{0pt}
    \caption{Details of the synthetic image generation process. \{class\} corresponds to a class name in the dataset.}
    \label{table:prompts}
    \centering
    \begin{tabular}{c|c|c|c}
        Dataset & Prompt & $1\times$ Generation Cost (hrs) & $2\times$ Generation Cost (hrs) \bigstrut[b] \\
        \hline
        MIT-67 & ``the inside of a \{class\}'' & 18.355 (80 imgs/class) & 36.711 (160 imgs/class) \bigstrut[t] \\
        DTD & ``\{class\} texture'' & 6.438 (40 imgs/class) & 12.876 (80 imgs/class) \\
        Caltech-101 & ``a picture of a \{class\}'' & 29.399 (68 imgs/class) & 58.799 (136 imgs/class) \bigstrut[b] \\
    \end{tabular}
\end{table}

\subsection{Ablations on Contrastive Loss Design}
\label{sec:ablations}

Our ablation experiments were done in the process of designing the A/U loss formulation. We tested
whether the original KD loss \citep{hinton2015distilling} or the newer SRRL loss \citep{yang2021srrl} should
be used for the logit-based loss. In addition, we experimented with the design of the projection module $g(\cdot)$.
The teacher-student pairing was ResNet50-MobileNetV2. The results can be found in Table \ref{table:ablations}.

\begin{table}[H]
    \caption{Ablation on different projector architectures and logit-based losses for our A/U loss.}
    \label{table:ablations}
    \begin{center}
    \begin{tabular}{c c c c c c c c}
    Projector & Logit Loss & CIFAR-10 & CIFAR-100 & MIT-67 & CUB-2011 & Caltech-101 & DTD \bigstrut[b] \\
    \hline
    MoCoV2 & SRRL & \textbf{93.25} & 73.29 & 65.82 & 63.03 & 87.23 & 49.63 \bigstrut[t] \\
    MoCoV2 & KD & 92.66 & \textbf{75.8} & 67.09 & 67.45 & \textbf{87.74} & \textbf{52.71} \\
    Linear & SRRL & 93.18 & 73.16 & 66.19 & 62.08 & 86.95 & 50.74 \\
    Linear & KD & 92.81 & \textbf{75.8} & \textbf{67.54} & \textbf{68.16} & 87.51 & 52.07\\
    \end{tabular}
    \end{center}
\end{table}

\begin{figure}[H]
    \centering
    \includegraphics[width=0.25\textwidth]{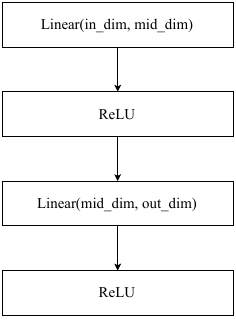}
    \caption{Our modified design of the module that projects the student features to the teacher dimensions in SRRL. mid\_dim is the average
    of in\_dim and out\_dim.}
    \label{fig:srrl_connector}
\end{figure}

\paragraph{$\mathcal{L}_{\text{KD}}$ vs. SRRL Loss}
In SRRL \citep{yang2021srrl}, instead of comparing the student and teacher logits, the student features are first passed through
a connector module (Figure \ref{fig:srrl_connector}) that lifts them to the teacher's feature space then passed through the teacher's classifier. Then,
these logits are compared. When comparing using this or $\mathcal{L}_{\text{KD}}$ for the logit-based loss, we found that
$\mathcal{L}_{\text{KD}}$ worked better.

\begin{figure}[H]
    \centering
    \includegraphics[width=0.25\textwidth]{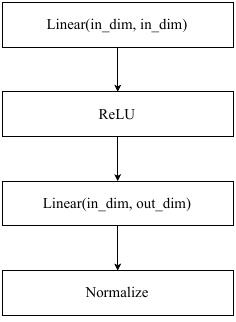}
    \caption{The design of the projector $g(\cdot)$ in A/U and CRD++.}
    \label{fig:projector}
\end{figure}

\paragraph{Projector Design}
In CRD \citep{tian2019crd}, the projection module to $\R^d$ is a linear matrix multiplication. Since its publication, improvements
on the contrastive learning regime have been proposed, one of which is the use of a \emph{deeper} projector \citep{chen2020mocov2,
chen2020simclr}.
Thus, we also compared using the 2-layer MLP architecture found in MoCoV2 \citep{chen2020mocov2} vs. a linear one. The exact design
can be found in Figure \ref{fig:projector}. For our A/U based loss,
we found that both the deeper and linear one achieved good performance but chose the deeper one to show the flexibility of our design.
Our source code provides different options for the choice of the projector.
Substituting the deeper projector in CRD exactly describes CRD++.

\subsection{Experiment Details}
\label{sec:experiment_details}
For training, linear probing, and distilling, we used the AdamW optimizer \citep{loshchilov2018adamw} with the default parameters: 
1e-3 initial learning rate, beta = (0.9, 0.999), epsilon=1e-8 and a weight decay of 0.01. 
No additional learning rate schedulers were used. All experiments were done for 250 epochs with a random seed of 9.
Inputs were centered, resized to 224x224, and randomly flipped horizontally. Pre-trained models were taken from PyTorch's model hub
\footnote{\url{https://pytorch.org/vision/stable/models.html}}.

\subsection{Dataset Details}
\label{sec:dataset_details}
We used the Indoor Scene Recognition dataset (MIT-67) \cite{quattoni2009mit_indoor}, 
the Caltech-UCSD Birds-200-2011 dataset (CUB-2011) \cite{Wah2011cub_2011}, the Describable Textures Dataset (DTD) \cite{cimpoi14dtd},
and CIFAR-10/100 \cite{krizhevsky2009cifar}.
The first 4 are considered data-limited. MIT-67 has 67 classes with 5360 images in the train set.
CUB-2011 has 200 classes with 5994 images in its train set. Caltech-101 has 6907 train images across 101 classes, though some classes have many more
(hundreds) of images while others have as low as 30. DTD has a train set of size 1880 for 47 classes. Lastly, CIFAR-10 and CIFAR-100
are standard
vision datasets with 50,000 training images across 10 and 100 classes, respectively.

\end{document}